\documentclass[sigconf]{acmart}
\AtBeginDocument{%
  \providecommand\BibTeX{{%
    \normalfont B\kern-0.5em{\scshape i\kern-0.25em b}\kern-0.8em\TeX}}}

\copyrightyear{2021}
\acmYear{2021}
\setcopyright{acmcopyright}\acmConference[SIGIR '21]{Proceedings of the 44th International ACM SIGIR Conference on Research and Development in Information Retrieval}{July 11--15, 2021}{Virtual Event, Canada}
\acmBooktitle{Proceedings of the 44th International ACM SIGIR Conference on Research and Development in Information Retrieval (SIGIR '21), July 11--15, 2021, Virtual Event, Canada}
\acmPrice{15.00}
\acmDOI{10.1145/3404835.3462785}
\acmISBN{978-1-4503-8037-9/21/07}

\usepackage[export]{adjustbox}
\usepackage{multirow}
\usepackage{amsmath}
 \usepackage{balance}
\begin{document}
\fancyhead{} 
\title{FedNLP: An interpretable NLP System to Decode Federal Reserve Communications}

\author{Jean Lee}
\orcid{0000-0002-7457-028X}
\affiliation{%
  \institution{The University of Sydney}
  \city{Sydney}
  \state{NSW}
  \country{Australia}
  \postcode{2006}
}
\email{jean.lee@sydney.edu.au}

\author{Hoyoul Luis Youn}
\affiliation{%
  \institution{KPMG Australia}
  \city{Sydney}
  \state{NSW}
  \country{Australia}
}
\email{yhy0215@gmail.com}

\author{Nicholas Stevens}
\affiliation{%
  \institution{NRS Technology}
  \city{Sydney}
  \state{NSW}
  \country{Australia}
}
\email{nicholas.robert.stevens@gmail.com}

\author{Josiah Poon}
\affiliation{%
  \institution{The University of Sydney}
  \city{Sydney}
  \state{NSW}
  \country{Australia}
}
\email{josiah.poon@sydney.edu.au}

\author{Soyeon Caren Han}
\affiliation{%
  \institution{The University of Sydney}
  \city{Sydney}
  \state{NSW}
  \country{Australia}
}
\email{caren.han@sydney.edu.au}
\authornote{Corresponding Author (caren.han@sydney.edu.au)}

\renewcommand{\shortauthors}{J. Lee et al.}

\begin{abstract}
The Federal Reserve System (the Fed) plays a significant role in affecting monetary policy and financial conditions worldwide. Although it is important to analyse the Fed’s communications to extract useful information, it is generally long-form and complex due to the ambiguous and esoteric nature of content. In this paper, we present FedNLP, an interpretable multi-component Natural Language Processing (NLP) system to decode Federal Reserve communications. This system is designed for end-users to explore how NLP techniques can assist their holistic understanding of the Fed’s communications with NO coding. Behind the scenes, FedNLP uses multiple NLP models from traditional machine learning algorithms to deep neural network architectures in each downstream task. The demonstration shows multiple results at once including sentiment analysis, summary of the document, prediction of the Federal Funds Rate movement and visualization for interpreting the prediction model's result. Our application system and demonstration are available at \url{https://fednlp.net}
\end{abstract}




\maketitle

\section{Introduction}
Over the years, the role of the U.S. Federal Reserve System (the Fed) has expanded due to changes in the monetary and financial conditions globally. The Fed's decisions have a chain effect on a broader range of economic factors like inflation, employment, value of currency, growth and loans \cite{javed2019impact}. Therefore, it is important to analyse the Fed’s communications that anchor and guide market expectations, however, it is generally long-form and complex due to the ambiguous and esoteric nature of content \cite{alan08, economictimes19}. Additionally, the Fed has increased their interest in research exploring the importance of Natural Language Processing (NLP) for macroeconomics \cite{fednlp20}. It is aligned with the remarkable progress in NLP that has seen the emergence of a massive number of model architectures (e.g. Transformers \cite{VaswaniSPUJGKP17}), and pre-trained models (e.g. BERT \cite{DevlinCLT19}, T5 \cite{raffel2020exploring}). Considering the fact that the Fed supervision carries vast amounts of unstructured data, the significant improvement in NLP research could assist their needs. However, there are no pilot studies to identify how NLP components could help end-users analyse Federal Reserve communications.

\begin{figure}[t]
  \centering
  \includegraphics[width=\linewidth]{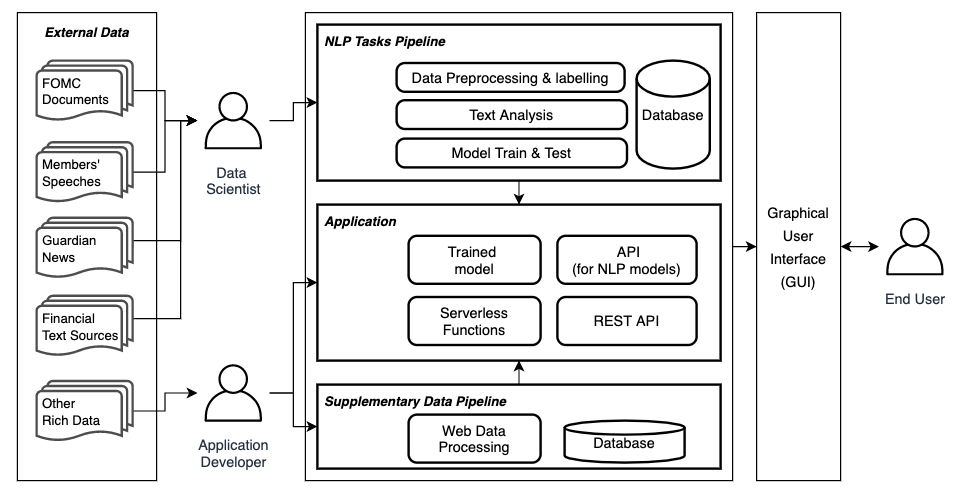}
  \caption{Functional System Flow of the FedNLP.}
  \label{fig:system_flow_user}
\end{figure}

In this paper, we present {\bfseries FedNLP}, an multi-component {\bfseries NLP} system that aims to decode {\bfseries Fed}eral Reserve communications with NO code. The system is designed for end-users to assist their holistic and intuitive understanding of the Fed’s communications through the use of multiple NLP components. We define that an end-user is the person who works within a broad range of business sectors such as finance and accounting, often reads economic and financial news, and has low to no programming skills. Our objectives are to reduce the gap between the advance of NLP technology and the needs for the use of NLP by building a "practical use" of NLP system. Inspired by recent research that combines language tools \cite{tenney2020language}, our system focuses on presenting multiple NLP components such as sentiment analysis \cite{LOUGHRAN2011}, prediction \cite{DevlinCLT19, yang2020finbert}, explanation \cite{Ribeiro0G16}, and summarization \cite{raffel2020exploring} in one application. Fig. \ref{fig:system_flow_user} shows the functional system flow that consists of NLP and application modules required to deliver a no-code system to an end-user. In each NLP task, multiple models are built and then the final models are selected for web applications. 

The main contributions of this paper are as follows: 
\begin{itemize}
\item 
To the best of our knowledge, we propose {\bfseries FedNLP}, the first interpretable multi-component NLP system for decoding Federal Reserve communications that assist end-users.
\item  
We implement the multiple NLP components and the associated algorithms from traditional machine learning models to pre-trained deep neural network models.
\item  
Our demonstration\footnote{Our demonstration system is available at https://fednlp.net.\\ Also, the screen recording of the demonstration can be found at \\ https://www.youtube.com/watch?v=Pn3OrWdzwws} facilitates the development of the system and shows the results from multiple NLP models at once. 
\item
Through our demo, an end-user can easily compare the different results from both general algorithms and the financial domain-specific algorithms.
\end{itemize}

\section{Background and Related Work}

\subsection{The Fed and FOMC}
The Federal Reserve (The Fed) controls the interest rate, specifically the Federal Funds Rate (FFR), in order to maximize employment rate and achieve stability in the prices of goods and services in the U.S. \cite{fedobj20}. Since the FFR indirectly impacts a very broad range of the economy, it is important to interpret the underlying factors which contribute to it. Historically, when the economy shows signs of weakness, like during the great recession or the Covid-19 pandemic, the Fed typically lowers rates. This decision is made by a committee within the Fed, called the Federal Open Market Committee (FOMC). As the FOMC has become more transparent with its communications \cite{schnidman2016fed} and has expanded their research interests to include the use of NLP \cite{fednlp20}, it is important to identify how NLP components could help end-users make better-informed decisions. In this research, we focused on FOMC members' communications such as reports, press releases and speeches as they also hold certain importance and insights for the market \cite{jung2016have, hayo2010federal}. 

\subsection{Visualisation Tools and Systems}
There has been some work on interactive analysis of understanding ML performance. Several systems have taken a black-box approach which does not rely on internal workings of a model, but is designed to let users examine inputs and outputs. Many are general-purpose and focusing on a visual inspection of model’s behavior on sample data, including ModelTracker \cite{AmershiCDLSS15}, Prospector \cite{KrausePN16}, Manifold \cite{ZhangWMLE19}, or What-If Tool \cite{WexlerPBWVW20}. For example, What-If Tool provides a rich support for intersectional analysis within dataset, tests hypothetical outcomes and focuses on ML fairness. 

In linguistic tasks, visualisation has shown to be useful tool for understanding deep neural networks such as LSTMVis \cite{StrobeltGPR18}, Seq2Seq-Vis \cite{StrobeltGBPPR19}, BertViz \cite{vig2019analyzing}, ExBERT \cite{hoover2020exbert}, or LIT \cite{tenney2020language}. Typical solutions include visualizing the internal structure or intermediate states of the model to enhance the understanding and interpretation, evaluating and analysing the performance of models or algorithms, and interactively improving the models at different development stages such as feature engineering or hyperparameter tuning through integration of domain knowledge. However, the focus of these tools has been restricted to developers, lacking the ability to deal with long documents or complex industry-specific documents. In this research, we design a visualization system for end-users to provide a holistic view of how NLP techniques analyse and interpret the Fed's communications.

\section{System Design and Implementation}
\begin{figure}[t]
  \centering
  \includegraphics[width=\linewidth]{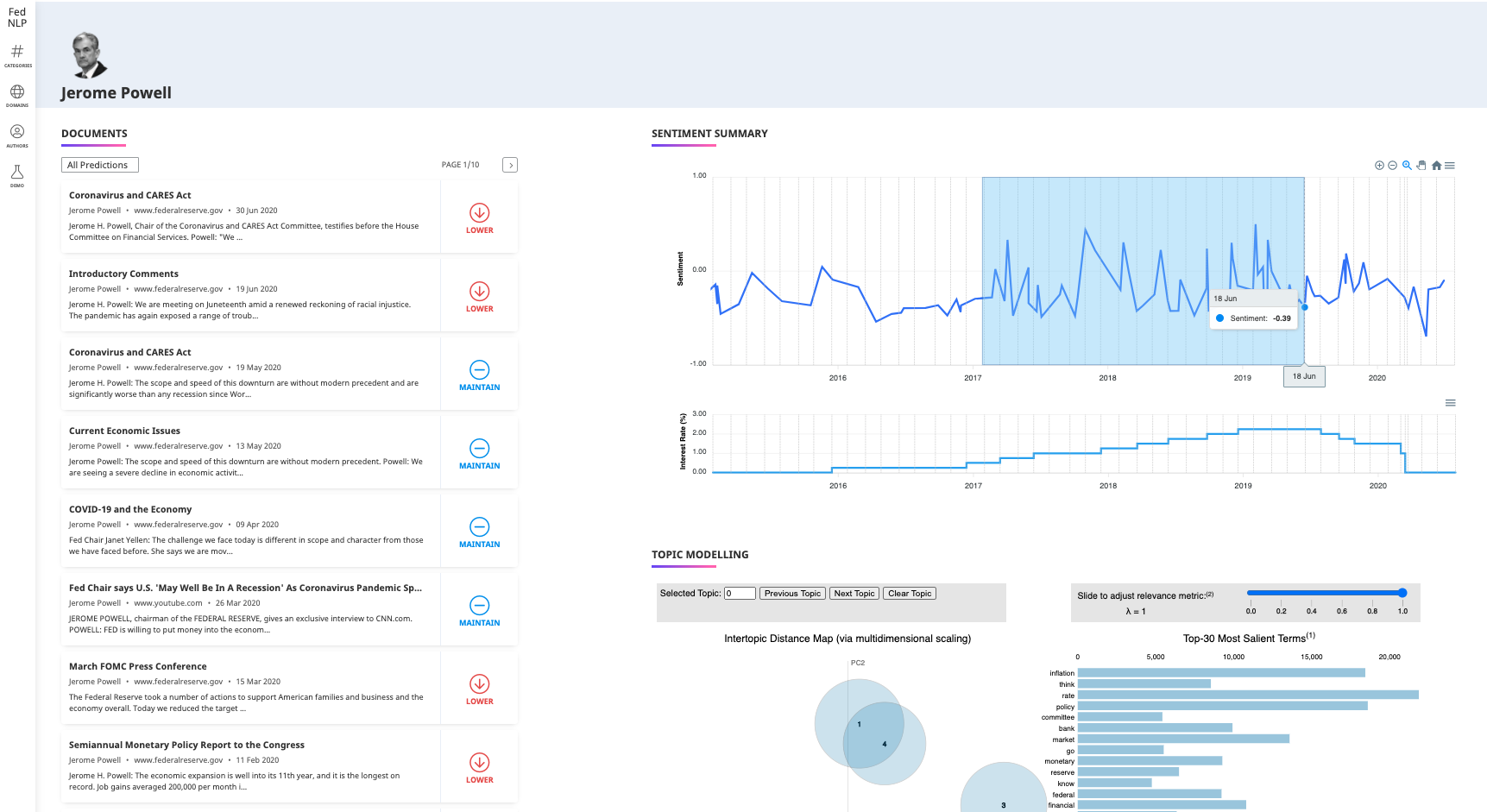}
  \caption{FedNLP Interface. The landing page of each FOMC member has {\itshape (top right)} a graph of the financial sentiment, a FFR graph over time, {\itshape (bottom right)} topic modelling, and a list of the document belongs to the member.}
  \label{fig:interface}
\end{figure}

Initially, we define an end-user and conduct preliminary focus group interviews by recruiting target end-users to identify system requirements. Through the in-depth interviews, we determine which functional components would be more useful, and design the proposed system and the functional system flow (Fig. \ref{fig:system_flow_user}). 
Additionally, we collect text data associated with the Fed’s communications from over 30 websites related to changes in target Federal Funds Rate (e.g. {\itshape lower, maintain, or raise}). With this real-world data, we implement eight widely used NLP components and algorithms (Table \ref{tab:fedNLP_components}). The sentiment analysis, prediction, explanation and summarization tasks provide a side-by-side comparison of generic and finance-specific algorithms and pose the question to end-users whether financial-specific algorithms are more capable of capturing the Fed’s communication than generic algorithms. 
Ultimately, the proposed system is deployed in a live environment, to provide a simple and easy to access environment for end-users (Fig. \ref{fig:interface}). 

\begin{table*}[ht]
\renewcommand{\arraystretch}{1.2}
\caption{Multiple language processing components and algorithms in the proposed FedNLP. "General" denotes general algorithms and "Financial" denotes the financial domain-specific algorithms.}
\label{tab:fedNLP_components}
\begin{adjustbox}{max width = \textwidth}
\begin{tabular}{lllcc}
\toprule
\multicolumn{1}{l}{\textbf{Component}} &
  \textbf{Algorithm} &
  \textbf{Description} &
  \multicolumn{1}{l}{\textbf{General}} &
  \multicolumn{1}{l}{\textbf{Financial}} \\ 
\midrule
\multirow{2}{*}{\textbf{Sentiment Analysis}} &
  TextBlob \cite{textblob17} &
  Returns polarity and subjectivity  using TextBlob for general settings. &
  v &
   \\ \cline{2-5} 
 &
  LM Sentiment \cite{LOUGHRAN2011} &
  Returns polarity and subjectivity using LM sentiment for financial settings.&
   &
  v \\ \hline
\textbf{Topic Modelling} &
  LDA \cite{Rehurek10softwareframework} &
  Visualizes term clusters and topics in HTML using LDA model. &
  v &
   \\ \hline
\multirow{2}{*}{\textbf{Prediction}} &
  XGBoost \cite{ChenG16} &
  Displays ML model predictions with explanation component. &
  v &
   \\ \cline{2-5} 
 &
  FinBERT \cite{yang2020finbert} &
  Displays model predictions using a fine-tuned FinBERT. &
   &
  v \\ \hline
\textbf{Explanation} &
  Lime \cite{Ribeiro0G16} &
  Visualizes top 10 highly-contributing features and highlights sentences. &
  v &
  v \\ \hline
\multirow{2}{*}{\textbf{Summarization}} &
  TextRank \cite{MihalceaT04} &
  Displays an extractive summarization using a graph-based ranking model &
  v &
   \\ \cline{2-5} 
 &
  T5 \cite{raffel2020exploring} &
  Displays an abstractive summarization using a fine-tuned T5. &
   &
  v \\ \hline
\textbf{Demonstration} &
  Decoupled APIs &
  Shows multi-components in one webpage that works with new input data. &
  v &
  v \\ 
\bottomrule
\end{tabular}
\end{adjustbox}
\end{table*}

\subsection{Sentiment Analysis}
In our system, we include sentiment analysis which shows the attitude or the emotion of the writer (e.g. positive, negative, or neutral). For the generic representation, we apply a TextBlob algorithm \cite{textblob17} that trains the data using {\itshape NLTK corpus} with naive bayes classifier. For the financial representation, we adopt the lexicon-based method for economic and financial documents, which was constructed by Loughran and McDonald (LM sentiment \cite{LOUGHRAN2011}). LM sentiment consists of financial word dictionaries appearing in corporate 10K/Q documents and earning calls.

\subsection{Prediction}
We more focus on predicting the direction of changes to the FFR by enabling the NLP models to be trained on the textual representations of the FOMC's decision. We conduct extensive experiments including traditional Machine Learning (ML), Neural Network (NN), and pre-trained models. Our approach is compared to following models: SVM, Linear SVC, Logistic Regression, Random Forest, XGBoost, CNN, a fine-tuned BERT, and a fine-tuned FinBERT. Among all the experiments on ML models, XGBoost with TF-IDF features achieve the highest test accuracy of 0.73 and weighted average F1 score of 0.66 while detecting all three classes. In Neural Network baseline models, the best setting on CNN and BERT base overfits and performs relatively poorly. A fine-tuned FinBERT achieves a test accuracy of 0.72 and weighted average F1 score of 0.65, however, detecting only maintain class which shows overfitting issues. In our system, we implement XGBoost with TF-IDF features and a fine-tuned FinBERT model based on the evaluation results. Due to space limits, we don't include the detailed results in this paper. 

\subsection{Explanation}
One of the questions from end-users was how to trust a prediction model and its results. The strength of text-based prediction models is that the model can provide an explanation that is easily understandable by people. In our research, we implement Local Interpretable Model-agnostic Explanations (LIME \cite{Ribeiro0G16}) that provides a visualisation by using the classifier's output to generate a linear surrogate model. The visualisation shows that the highly contributed words are different for each prediction model even though the prediction results are the same. A fine-tuned FinBERT captures more financial and economic jargon than any other model, however, is computationally expensive. In the system, we chose XGBoost with LIME as the explanation module because of computation costs. For the deployment, we further optimises the delivery of the explanation by isolating the result from the visualisation library (D3). The choice of XGBoost and delivery optimisation reduced the time to deliver a result from the explanation module from 10 minutes to 30 seconds.

\subsection{Summarization}
The summarization module provides key information about the document, giving users a simple way to quickly decide whether or not to read the full content. Automatic summarization gives a direct benefit because the Fed’s documents are often long and complex. Automatic summarization has been studied for decades and there are two types of categories - extractive and abstractive. Extractive summarization aims to capture most information with the least redundancy whereas abstractive summarization aims to generate new sentences to encapsulate maximum gist of the input document. In this research, we use TextRank \cite{MihalceaT04} and a fine-tuned T5 \cite{raffel2020exploring} for extractive and abstractive summarization, respectively.

\begin{figure*}[t]
  \centering
  \includegraphics[width=0.85\textwidth]{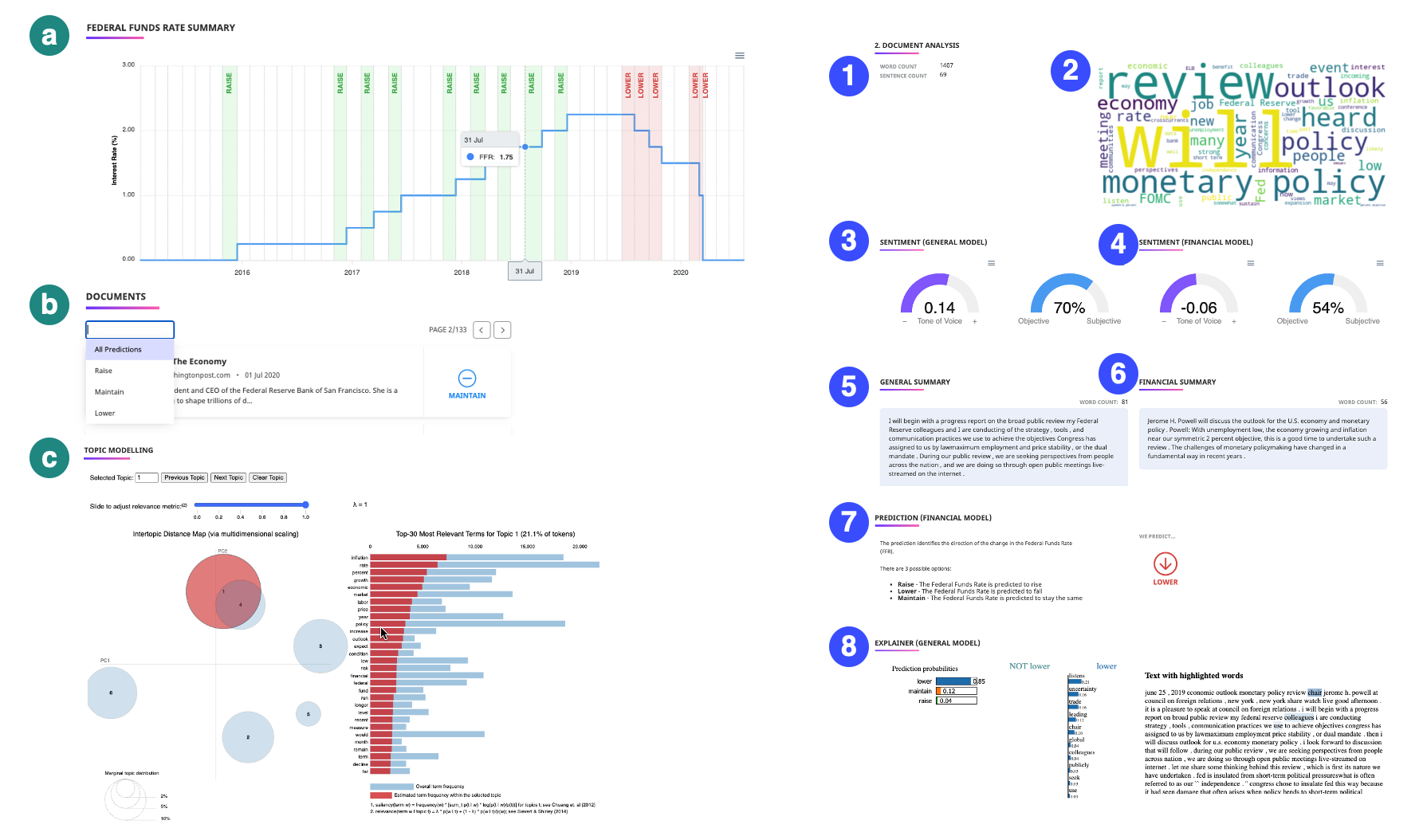}
  \caption{FedNLP System Components. {\itshape Left: Landing page components}  (a) Federal Funds Target Rate (Lower Bound) Graph and the Fed’s Decision (b) a List and Filtering Function for the Fed’s Documents (c) Topic Modelling Graph. {\itshape Right: Demo page components} (1) Word, Sentence Count  (2) WordCloud (3) General Sentiment (TextBlob) (4) Financial Sentiment (LM) (5) Summarization (General, TextRank) (6) Summarization (Financial, fine tuning T5) (7) Prediction (Financial, fine tuning FinBERT) (8) Explanation (General, XGBoost).}
  \label{fig:system_elements}
\end{figure*}

\section{Web Application}
In order to provide a simple no code experience, all components are delivered to end-users through a familiar interface - a web application (Fig \ref{tab:fedNLP_components}). 


\subsection{Graphic User Interface (GUI)}
The GUI is a lightweight web-based Angular application that provides a simple, intuitive interface for end-users to explore documents and the related predictions. Following Angular design principles, the application architecture consists of components (e.g. document listing and graphs), and page components for areas. Additionally, there are six data services that modularise access a single REST API or NLP API. These services make HTTP requests to an API endpoint and parse the response into JavaScript objects for use within components. The GUI is hosted on AWS, published to an S3 bucket and deployed as a static website using AWS CloudFront for distribution and AWS Route 53 for domain resolution.


\subsection{REST API}
The REST API delivers static content in JSON format to the GUI. The API consists of Node.js microservices deployed to AWS Lambda in a serverless configuration. The microservice is a simple data-retrieval script that 1) determines which object/s to retrieve based on path parameters, 2) makes the call to the AWS DynamoDB database containing the static content, 3) formats and emits the response into JSON data. In order to improve the application performance and data load time, the document data is split into two endpoints; {\itshape documents} contains lightweight data used in lists and section pages, and the {\itshape document-extensions} endpoint returns full document content used on the document view only. A simple Node.js script extracts and splits document and document-extension content from the NLP pipeline document data. A dedicated DynamoDB load script uploads this data into the corresponding table. While category, domain, and author content were manually sourced, it was transformed and uploaded with the same DynamoDB load script. 


\subsection{NLP API}
The NLP API is an endpoint for text analysis (WordCloud, sentiment analysis and topic modelling) and NLP tasks (prediction, explanation and summarization). The API itself consists of a simple Flask web server running inside a Docker container on an AWS EC2 server. The Docker environment initialises with all required libraries, including PyTorch and TensorFlow, as well as the configuration for Flask and network. On initialisation, the trained and packaged model from the NLP pipeline is downloaded from an AWS S3 bucket onto the server. The Flask API exposes one route for text analysis or each NLP task, handles routing, data input parsing, execution of the underlying code and formatting and emitting the response as JSON data.

\section{Conclusion}
We propose the {\bfseries FedNLP} system, which is designed to let end-users explore various NLP analyses and tasks to assist with decoding Federal Reserve communications. To the best of our knowledge, our system is the first of its kind to present the use of NLP in analysing many forms of Fed’s documents including post-meeting minutes, members’ speeches, and transcripts. The system enables end-users to experiment with custom input through an interactive demo which presents multiple NLP analysis results automatically through decoupled, productionized APIs. In practical use, FedNLP will be emphasized as a supplementary system to provide text analysis indicators from Federal Reserve communications and conduct further empirical studies.



\begin{acks}
We thank Gautam Radhakrishnan Ajit, Manisha Gupta, Renil Austin Mendez, and Lavanshu Agrawal for helping us to collect data and formulate the problem. We would also like to thank the anonymous reviewers and the NLP group at the University of Sydney for providing us with valuable comments. 
\end{acks}

\bibliographystyle{ACM-Reference-Format}
\balance
\bibliography{2_bibliography}


\begin{thebibliography}{27}


\ifx \showCODEN    \undefined \def \showCODEN     #1{\unskip}     \fi
\ifx \showDOI      \undefined \def \showDOI       #1{#1}\fi
\ifx \showISBNx    \undefined \def \showISBNx     #1{\unskip}     \fi
\ifx \showISBNxiii \undefined \def \showISBNxiii  #1{\unskip}     \fi
\ifx \showISSN     \undefined \def \showISSN      #1{\unskip}     \fi
\ifx \showLCCN     \undefined \def \showLCCN      #1{\unskip}     \fi
\ifx \shownote     \undefined \def \shownote      #1{#1}          \fi
\ifx \showarticletitle \undefined \def \showarticletitle #1{#1}   \fi
\ifx \showURL      \undefined \def \showURL       {\relax}        \fi
\providecommand\bibfield[2]{#2}
\providecommand\bibinfo[2]{#2}
\providecommand\natexlab[1]{#1}
\providecommand\showeprint[2][]{arXiv:#2}

\bibitem[\protect\citeauthoryear{Amershi, Chickering, Drucker, Lee, Simard, and
  Suh}{Amershi et~al\mbox{.}}{2015}]%
        {AmershiCDLSS15}
\bibfield{author}{\bibinfo{person}{Saleema Amershi}, \bibinfo{person}{Max
  Chickering}, \bibinfo{person}{Steven~M Drucker}, \bibinfo{person}{Bongshin
  Lee}, \bibinfo{person}{Patrice Simard}, {and} \bibinfo{person}{Jina Suh}.}
  \bibinfo{year}{2015}\natexlab{}.
\newblock \showarticletitle{Modeltracker: Redesigning performance analysis
  tools for machine learning}. In \bibinfo{booktitle}{\emph{Proceedings of the
  33rd Annual ACM Conference on Human Factors in Computing Systems}}.
  \bibinfo{publisher}{{ACM}}, \bibinfo{address}{New York, NY, USA},
  \bibinfo{pages}{337--346}.
\newblock


\bibitem[\protect\citeauthoryear{Blinder, Ehrmann, Fratzscher, De~Haan, and
  Jansen}{Blinder et~al\mbox{.}}{2008}]%
        {alan08}
\bibfield{author}{\bibinfo{person}{Alan~S Blinder}, \bibinfo{person}{Michael
  Ehrmann}, \bibinfo{person}{Marcel Fratzscher}, \bibinfo{person}{Jakob
  De~Haan}, {and} \bibinfo{person}{David-Jan Jansen}.}
  \bibinfo{year}{2008}\natexlab{}.
\newblock \showarticletitle{Central bank communication and monetary policy: A
  survey of theory and evidence}.
\newblock \bibinfo{journal}{\emph{Journal of economic literature}}
  \bibinfo{volume}{46}, \bibinfo{number}{4} (\bibinfo{year}{2008}),
  \bibinfo{pages}{910--45}.
\newblock


\bibitem[\protect\citeauthoryear{Board of Governors of~the Federal
  Reserve~System}{Board of Governors of~the Federal Reserve~System}{2020a}]%
        {fedobj20}
\bibfield{author}{\bibinfo{person}{TheFed Board of Governors of~the Federal
  Reserve~System}.} \bibinfo{year}{2020}\natexlab{a}.
\newblock \bibinfo{title}{What economic goals does the Federal Reserve seek to
  achieve through its monetary policy?}
\newblock
\newblock
\urldef\tempurl%
\url{https://www.federalreserve.gov/faqs/what-economic-goals-does-federal-reserve-seek-to-achieve-through-monetary-policy.htm}
\showURL{%
Retrieved October 10, 2020 from \tempurl}


\bibitem[\protect\citeauthoryear{Board of Governors of~the Federal
  Reserve~System}{Board of Governors of~the Federal Reserve~System}{2020b}]%
        {fednlp20}
\bibfield{author}{\bibinfo{person}{TheFed Board of Governors of~the Federal
  Reserve~System}.} \bibinfo{year}{2020}\natexlab{b}.
\newblock \bibinfo{title}{Workshop on the Use of Natural Language Processing in
  Supervision}.
\newblock
\newblock
\urldef\tempurl%
\url{https://www.federalreserve.gov/conferences/workshop-on-the-use-of-natural-language-in-supervision.htm}
\showURL{%
Retrieved October 16, 2020 from \tempurl}


\bibitem[\protect\citeauthoryear{Chen and Guestrin}{Chen and Guestrin}{2016}]%
        {ChenG16}
\bibfield{author}{\bibinfo{person}{Tianqi Chen} {and} \bibinfo{person}{Carlos
  Guestrin}.} \bibinfo{year}{2016}\natexlab{}.
\newblock \showarticletitle{Xgboost: A scalable tree boosting system}. In
  \bibinfo{booktitle}{\emph{Proceedings of the 22nd acm sigkdd international
  conference on knowledge discovery and data mining}}.
  \bibinfo{publisher}{{ACM}}, \bibinfo{address}{New York, NY, USA},
  \bibinfo{pages}{785--794}.
\newblock


\bibitem[\protect\citeauthoryear{Devlin, Chang, Lee, and Toutanova}{Devlin
  et~al\mbox{.}}{2019}]%
        {DevlinCLT19}
\bibfield{author}{\bibinfo{person}{Jacob Devlin}, \bibinfo{person}{Ming{-}Wei
  Chang}, \bibinfo{person}{Kenton Lee}, {and} \bibinfo{person}{Kristina
  Toutanova}.} \bibinfo{year}{2019}\natexlab{}.
\newblock \showarticletitle{{BERT:} Pre-training of Deep Bidirectional
  Transformers for Language Understanding}. In
  \bibinfo{booktitle}{\emph{Proceedings of the 2019 Conference of the North
  American Chapter of the Association for Computational Linguistics: Human
  Language Technologies, {NAACL-HLT} 2019}}. \bibinfo{publisher}{Association
  for Computational Linguistics}, \bibinfo{address}{Minneapolis, Minnesota,
  USA}, \bibinfo{pages}{4171--4186}.
\newblock


\bibitem[\protect\citeauthoryear{Economic~Times}{Economic~Times}{2019}]%
        {economictimes19}
\bibfield{author}{\bibinfo{person}{The Economic~Times}.}
  \bibinfo{year}{2019}\natexlab{}.
\newblock \bibinfo{title}{Decoding central bankers’ language}.
\newblock
\newblock
\urldef\tempurl%
\url{https://economictimes.indiatimes.com/markets/stocks/news/decoding-central-bankers-language/articleshow/69952364.cms}
\showURL{%
Retrieved October 10, 2020 from \tempurl}


\bibitem[\protect\citeauthoryear{Hayo and Neuenkirch}{Hayo and
  Neuenkirch}{2010}]%
        {hayo2010federal}
\bibfield{author}{\bibinfo{person}{Bernd Hayo} {and} \bibinfo{person}{Matthias
  Neuenkirch}.} \bibinfo{year}{2010}\natexlab{}.
\newblock \showarticletitle{Do Federal Reserve communications help predict
  federal funds target rate decisions?}
\newblock \bibinfo{journal}{\emph{Journal of Macroeconomics}}
  \bibinfo{volume}{32}, \bibinfo{number}{4} (\bibinfo{year}{2010}),
  \bibinfo{pages}{1014--1024}.
\newblock


\bibitem[\protect\citeauthoryear{Hoover, Strobelt, and Gehrmann}{Hoover
  et~al\mbox{.}}{2020}]%
        {hoover2020exbert}
\bibfield{author}{\bibinfo{person}{Benjamin Hoover}, \bibinfo{person}{Hendrik
  Strobelt}, {and} \bibinfo{person}{Sebastian Gehrmann}.}
  \bibinfo{year}{2020}\natexlab{}.
\newblock \showarticletitle{exBERT: A Visual Analysis Tool to Explore Learned
  Representations in Transformer Models}. In
  \bibinfo{booktitle}{\emph{Proceedings of the 58th Annual Meeting of the
  Association for Computational Linguistics: System Demonstrations}}.
  \bibinfo{publisher}{Association for Computational Linguistics},
  \bibinfo{address}{Online}, \bibinfo{pages}{187--196}.
\newblock


\bibitem[\protect\citeauthoryear{Javed, Atallah~Aldalaien, Husain, and
  Shahfaraz~Khan}{Javed et~al\mbox{.}}{2019}]%
        {javed2019impact}
\bibfield{author}{\bibinfo{person}{Sarfaraz Javed}, \bibinfo{person}{B
  Atallah~Aldalaien}, \bibinfo{person}{Uvesh Husain}, {and} \bibinfo{person}{M
  Shahfaraz~Khan}.} \bibinfo{year}{2019}\natexlab{}.
\newblock \showarticletitle{Impact of Federal Funds Rate on Monthly Stocks
  Return of United States of America}.
\newblock \bibinfo{journal}{\emph{International Journal of Business and
  Management}} \bibinfo{volume}{14}, \bibinfo{number}{9}
  (\bibinfo{year}{2019}), \bibinfo{pages}{105}.
\newblock


\bibitem[\protect\citeauthoryear{Jung}{Jung}{2016}]%
        {jung2016have}
\bibfield{author}{\bibinfo{person}{Alexander Jung}.}
  \bibinfo{year}{2016}\natexlab{}.
\newblock \showarticletitle{Have minutes helped to predict fed funds rate
  changes?}
\newblock \bibinfo{journal}{\emph{Journal of Macroeconomics}}
  \bibinfo{volume}{49} (\bibinfo{year}{2016}), \bibinfo{pages}{18--32}.
\newblock


\bibitem[\protect\citeauthoryear{Krause, Perer, and Ng}{Krause
  et~al\mbox{.}}{2016}]%
        {KrausePN16}
\bibfield{author}{\bibinfo{person}{Josua Krause}, \bibinfo{person}{Adam Perer},
  {and} \bibinfo{person}{Kenney Ng}.} \bibinfo{year}{2016}\natexlab{}.
\newblock \showarticletitle{Interacting with predictions: Visual inspection of
  black-box machine learning models}. In \bibinfo{booktitle}{\emph{Proceedings
  of the 2016 CHI conference on human factors in computing systems}}.
  \bibinfo{publisher}{{ACM}}, \bibinfo{address}{San Jose, CA, USA},
  \bibinfo{pages}{5686--5697}.
\newblock


\bibitem[\protect\citeauthoryear{Loria}{Loria}{2017}]%
        {textblob17}
\bibfield{author}{\bibinfo{person}{Steven Loria}.}
  \bibinfo{year}{2017}\natexlab{}.
\newblock \bibinfo{title}{TextBlob: Simplified Text Processing}.
\newblock
\newblock
\urldef\tempurl%
\url{https://textblob.readthedocs.io/en/dev/}
\showURL{%
Retrieved October 10, 2020 from \tempurl}


\bibitem[\protect\citeauthoryear{Loughran and McDonald}{Loughran and
  McDonald}{2011}]%
        {LOUGHRAN2011}
\bibfield{author}{\bibinfo{person}{Tim Loughran} {and} \bibinfo{person}{Bill
  McDonald}.} \bibinfo{year}{2011}\natexlab{}.
\newblock \showarticletitle{When is a liability not a liability? Textual
  analysis, dictionaries, and 10-Ks}.
\newblock \bibinfo{journal}{\emph{The Journal of finance}}
  \bibinfo{volume}{66}, \bibinfo{number}{1} (\bibinfo{year}{2011}),
  \bibinfo{pages}{35--65}.
\newblock


\bibitem[\protect\citeauthoryear{Mihalcea and Tarau}{Mihalcea and
  Tarau}{2004}]%
        {MihalceaT04}
\bibfield{author}{\bibinfo{person}{Rada Mihalcea} {and} \bibinfo{person}{Paul
  Tarau}.} \bibinfo{year}{2004}\natexlab{}.
\newblock \showarticletitle{Textrank: Bringing order into text}. In
  \bibinfo{booktitle}{\emph{Proceedings of the 2004 conference on empirical
  methods in natural language processing}}. \bibinfo{publisher}{{ACL}},
  \bibinfo{address}{Barcelona, Spain}, \bibinfo{pages}{404--411}.
\newblock


\bibitem[\protect\citeauthoryear{Raffel, Shazeer, Roberts, Lee, Narang, Matena,
  Zhou, Li, and Liu}{Raffel et~al\mbox{.}}{2020}]%
        {raffel2020exploring}
\bibfield{author}{\bibinfo{person}{Colin Raffel}, \bibinfo{person}{Noam
  Shazeer}, \bibinfo{person}{Adam Roberts}, \bibinfo{person}{Katherine Lee},
  \bibinfo{person}{Sharan Narang}, \bibinfo{person}{Michael Matena},
  \bibinfo{person}{Yanqi Zhou}, \bibinfo{person}{Wei Li}, {and}
  \bibinfo{person}{Peter~J Liu}.} \bibinfo{year}{2020}\natexlab{}.
\newblock \showarticletitle{Exploring the Limits of Transfer Learning with a
  Unified Text-to-Text Transformer}.
\newblock \bibinfo{journal}{\emph{Journal of Machine Learning Research}}
  \bibinfo{volume}{21} (\bibinfo{year}{2020}), \bibinfo{pages}{1--67}.
\newblock


\bibitem[\protect\citeauthoryear{Rehurek and Sojka}{Rehurek and Sojka}{2010}]%
        {Rehurek10softwareframework}
\bibfield{author}{\bibinfo{person}{Radim Rehurek} {and} \bibinfo{person}{Petr
  Sojka}.} \bibinfo{year}{2010}\natexlab{}.
\newblock \showarticletitle{Software framework for topic modelling with large
  corpora}. In \bibinfo{booktitle}{\emph{In Proceedings of the LREC 2010
  workshop on new challenges for NLP frameworks}}. Citeseer,
  \bibinfo{publisher}{European Language Resources Association (ELRA)},
  \bibinfo{address}{Valletta, Malta}, \bibinfo{pages}{46--50}.
\newblock


\bibitem[\protect\citeauthoryear{Ribeiro, Singh, and Guestrin}{Ribeiro
  et~al\mbox{.}}{2016}]%
        {Ribeiro0G16}
\bibfield{author}{\bibinfo{person}{Marco~T{\'{u}}lio Ribeiro},
  \bibinfo{person}{Sameer Singh}, {and} \bibinfo{person}{Carlos Guestrin}.}
  \bibinfo{year}{2016}\natexlab{}.
\newblock \showarticletitle{"Why Should {I} Trust You?": Explaining the
  Predictions of Any Classifier}. In \bibinfo{booktitle}{\emph{Proceedings of
  the 22nd {ACM} {SIGKDD} International Conference on Knowledge Discovery and
  Data Mining}}. \bibinfo{publisher}{{ACM}}, \bibinfo{address}{San Francisco,
  CA, USA}, \bibinfo{pages}{1135--1144}.
\newblock


\bibitem[\protect\citeauthoryear{Schnidman and MacMillan}{Schnidman and
  MacMillan}{2016}]%
        {schnidman2016fed}
\bibfield{author}{\bibinfo{person}{Evan~A Schnidman} {and}
  \bibinfo{person}{William~D MacMillan}.} \bibinfo{year}{2016}\natexlab{}.
\newblock \bibinfo{booktitle}{\emph{How the Fed Moves Markets: Central Bank
  Analysis for the Modern Era}}.
\newblock \bibinfo{publisher}{Springer}, \bibinfo{address}{New York, USA}.
\newblock


\bibitem[\protect\citeauthoryear{Strobelt, Gehrmann, Behrisch, Perer, Pfister,
  and Rush}{Strobelt et~al\mbox{.}}{2018}]%
        {StrobeltGBPPR19}
\bibfield{author}{\bibinfo{person}{Hendrik Strobelt},
  \bibinfo{person}{Sebastian Gehrmann}, \bibinfo{person}{Michael Behrisch},
  \bibinfo{person}{Adam Perer}, \bibinfo{person}{Hanspeter Pfister}, {and}
  \bibinfo{person}{Alexander~M Rush}.} \bibinfo{year}{2018}\natexlab{}.
\newblock \showarticletitle{S eq 2s eq-v is: A visual debugging tool for
  sequence-to-sequence models}.
\newblock \bibinfo{journal}{\emph{IEEE transactions on visualization and
  computer graphics}} \bibinfo{volume}{25}, \bibinfo{number}{1}
  (\bibinfo{year}{2018}), \bibinfo{pages}{353--363}.
\newblock


\bibitem[\protect\citeauthoryear{Strobelt, Gehrmann, Pfister, and
  Rush}{Strobelt et~al\mbox{.}}{2017}]%
        {StrobeltGPR18}
\bibfield{author}{\bibinfo{person}{Hendrik Strobelt},
  \bibinfo{person}{Sebastian Gehrmann}, \bibinfo{person}{Hanspeter Pfister},
  {and} \bibinfo{person}{Alexander~M Rush}.} \bibinfo{year}{2017}\natexlab{}.
\newblock \showarticletitle{Lstmvis: A tool for visual analysis of hidden state
  dynamics in recurrent neural networks}.
\newblock \bibinfo{journal}{\emph{IEEE transactions on visualization and
  computer graphics}} \bibinfo{volume}{24}, \bibinfo{number}{1}
  (\bibinfo{year}{2017}), \bibinfo{pages}{667--676}.
\newblock


\bibitem[\protect\citeauthoryear{Tenney, Wexler, Bastings, Bolukbasi, Coenen,
  Gehrmann, Jiang, Pushkarna, Radebaugh, Reif, et~al\mbox{.}}{Tenney
  et~al\mbox{.}}{2020}]%
        {tenney2020language}
\bibfield{author}{\bibinfo{person}{Ian Tenney}, \bibinfo{person}{James Wexler},
  \bibinfo{person}{Jasmijn Bastings}, \bibinfo{person}{Tolga Bolukbasi},
  \bibinfo{person}{Andy Coenen}, \bibinfo{person}{Sebastian Gehrmann},
  \bibinfo{person}{Ellen Jiang}, \bibinfo{person}{Mahima Pushkarna},
  \bibinfo{person}{Carey Radebaugh}, \bibinfo{person}{Emily Reif},
  {et~al\mbox{.}}} \bibinfo{year}{2020}\natexlab{}.
\newblock \showarticletitle{The Language Interpretability Tool: Extensible,
  Interactive Visualizations and Analysis for NLP Models}. In
  \bibinfo{booktitle}{\emph{Proceedings of the 2020 Conference on Empirical
  Methods in Natural Language Processing: System Demonstrations}}.
  \bibinfo{publisher}{Association for Computational Linguistics},
  \bibinfo{address}{Online}, \bibinfo{pages}{107--118}.
\newblock


\bibitem[\protect\citeauthoryear{Vaswani, Shazeer, Parmar, Uszkoreit, Jones,
  Gomez, Kaiser, and Polosukhin}{Vaswani et~al\mbox{.}}{2017}]%
        {VaswaniSPUJGKP17}
\bibfield{author}{\bibinfo{person}{Ashish Vaswani}, \bibinfo{person}{Noam
  Shazeer}, \bibinfo{person}{Niki Parmar}, \bibinfo{person}{Jakob Uszkoreit},
  \bibinfo{person}{Llion Jones}, \bibinfo{person}{Aidan~N. Gomez},
  \bibinfo{person}{Lukasz Kaiser}, {and} \bibinfo{person}{Illia Polosukhin}.}
  \bibinfo{year}{2017}\natexlab{}.
\newblock \showarticletitle{Attention is All you Need}. In
  \bibinfo{booktitle}{\emph{Advances in Neural Information Processing Systems
  30: Annual Conference on Neural Information Processing Systems 2017}}.
  \bibinfo{publisher}{Curran Associates, Inc.}, \bibinfo{address}{Long Beach,
  CA, USA}, \bibinfo{pages}{5998--6008}.
\newblock


\bibitem[\protect\citeauthoryear{Vig and Belinkov}{Vig and Belinkov}{2019}]%
        {vig2019analyzing}
\bibfield{author}{\bibinfo{person}{Jesse Vig} {and} \bibinfo{person}{Yonatan
  Belinkov}.} \bibinfo{year}{2019}\natexlab{}.
\newblock \showarticletitle{Analyzing the Structure of Attention in a
  Transformer Language Model}. In \bibinfo{booktitle}{\emph{Proceedings of the
  2019 ACL Workshop BlackboxNLP: Analyzing and Interpreting Neural Networks for
  NLP}}. \bibinfo{publisher}{{ACL}}, \bibinfo{address}{Florence, Italy},
  \bibinfo{pages}{63--76}.
\newblock


\bibitem[\protect\citeauthoryear{Wexler, Pushkarna, Bolukbasi, Wattenberg,
  Vi{\'e}gas, and Wilson}{Wexler et~al\mbox{.}}{2019}]%
        {WexlerPBWVW20}
\bibfield{author}{\bibinfo{person}{James Wexler}, \bibinfo{person}{Mahima
  Pushkarna}, \bibinfo{person}{Tolga Bolukbasi}, \bibinfo{person}{Martin
  Wattenberg}, \bibinfo{person}{Fernanda Vi{\'e}gas}, {and}
  \bibinfo{person}{Jimbo Wilson}.} \bibinfo{year}{2019}\natexlab{}.
\newblock \showarticletitle{The what-if tool: Interactive probing of machine
  learning models}.
\newblock \bibinfo{journal}{\emph{IEEE transactions on visualization and
  computer graphics}} \bibinfo{volume}{26}, \bibinfo{number}{1}
  (\bibinfo{year}{2019}), \bibinfo{pages}{56--65}.
\newblock


\bibitem[\protect\citeauthoryear{Yang, UY, and Huang}{Yang
  et~al\mbox{.}}{2020}]%
        {yang2020finbert}
\bibfield{author}{\bibinfo{person}{Yi Yang}, \bibinfo{person}{Mark
  Christopher~Siy UY}, {and} \bibinfo{person}{Allen Huang}.}
  \bibinfo{year}{2020}\natexlab{}.
\newblock \showarticletitle{Finbert: A pretrained language model for financial
  communications}.
\newblock \bibinfo{journal}{\emph{arXiv preprint arXiv:2006.08097}}
  (\bibinfo{year}{2020}).
\newblock


\bibitem[\protect\citeauthoryear{Zhang, Wang, Molino, Li, and Ebert}{Zhang
  et~al\mbox{.}}{2018}]%
        {ZhangWMLE19}
\bibfield{author}{\bibinfo{person}{Jiawei Zhang}, \bibinfo{person}{Yang Wang},
  \bibinfo{person}{Piero Molino}, \bibinfo{person}{Lezhi Li}, {and}
  \bibinfo{person}{David~S Ebert}.} \bibinfo{year}{2018}\natexlab{}.
\newblock \showarticletitle{Manifold: A model-agnostic framework for
  interpretation and diagnosis of machine learning models}.
\newblock \bibinfo{journal}{\emph{IEEE transactions on visualization and
  computer graphics}} \bibinfo{volume}{25}, \bibinfo{number}{1}
  (\bibinfo{year}{2018}), \bibinfo{pages}{364--373}.
\newblock


\end{thebibliography}


\end{document}